\pdfoutput=1

\documentclass[11pt]{article}

\usepackage[]{ACL2023}

\usepackage{times}
\usepackage{latexsym}

\usepackage[T1]{fontenc}

\usepackage[utf8]{inputenc}

\usepackage{microtype}

\usepackage{inconsolata}
\usepackage{epsfig}
\usepackage{subcaption}
\usepackage{calc}
\usepackage{amssymb}
\usepackage{amstext}
\usepackage{amsmath}
\usepackage{amsthm}
\usepackage{multicol}
\usepackage{pslatex}
\usepackage{algorithm2e}
\usepackage[bottom]{footmisc}
\usepackage{multirow, xcolor}
\usepackage{tabularx, color}
\usepackage{appendix}
\usepackage{longtable}
\usepackage{caption}
\usepackage[absolute,overlay]{textpos}
\usepackage{makecell}  
\usepackage{graphicx} 
\usepackage{booktabs}
\usepackage{geometry} 
\usepackage{array} 
\usepackage[utf8]{inputenc}
\usepackage{adjustbox}
\usepackage[T1]{fontenc}

\DeclareUnicodeCharacter{2217}{*}

\definecolor{pastelgreen}{gray}{0.8} 
\definecolor{pastelred}{gray}{0.5}   

\title{The Anatomy of Speech Persuasion: Linguistic Shifts in LLM-Modified Speeches}

\author{
Alisa Barkar \\
  \small Telecom Paris \\
  \small\texttt{alisa.barkar@telecom-paris.fr}
  \And
  Mathieu Chollet \\
  \small University of Glasgow \\ \small\& IMT Atlantique \\
  \small\texttt{mathieu.chollet@glasgow.ac.uk}
  \And
  Matthieu Labeau \\
  \small Telecom Paris \\
  \small\texttt{matthieu.labeau@telecom-paris.fr}
  \AND
  Beatrice Biancardi \\
  \small CESI LINEACT \\
  \small\texttt{bbiancardi@cesi.fr}
  \And
  Chloé Clavel \\
  \small INRIA ALMAnaCH \\
  \small\texttt{chloe.clavel@inria.fr}
}

\begin{document}
\maketitle

\begin{abstract}
This study examines how large language models understand the concept of persuasiveness in public speaking by modifying speech transcripts from PhD candidates in the ``Ma Th\`ese en 180 Secondes'' competition, using the 3MT\_French dataset. Our contributions include a novel methodology and an interpretable textual feature set integrating rhetorical devices and discourse markers. We prompt GPT-4o to enhance or diminish persuasiveness and analyze linguistic shifts between original and generated speech in terms of the new features. Results indicate that GPT-4o applies systematic stylistic modifications rather than optimizing persuasiveness in a human-like manner. Notably, it manipulates emotional lexicon and syntactic structures—such as interrogative and exclamatory clauses—to amplify rhetorical impact.
\end{abstract}

\section{Introduction}

Public speaking is a crucial skill, yet its analysis has predominantly focused on audio and visual modalities~\cite{Nguyen2012OnlineFS,Chen2015UtilizingMC}, leaving the role of textual content in shaping persuasiveness largely understudied especially in languages other than English. Meanwhile, recent advancements in Large Language Models (LLMs) have led to their integration into automated public speaking evaluation systems, enabling expert-like multilingual feedback on performance, specifically, on text\footnote{\scriptsize Yoodli: \url{https://app.yoodli.ai/}, GPT-based: \url{https://bit.ly/48eqllR} or \url{https://bit.ly/3YeKnYy}, Read: \url{https://www.read.ai/}}. However, while studies on recommendation systems~\cite{Lubos2024LLMgeneratedEF} suggest that users find LLM-generated feedback clear and comprehensive, prior research on LLM evaluation ability in text-related tasks shows weaknesses with engagement, coherence, persuasiveness, originality and story elaboration text evaluation~\cite{chhun2024languagemodelsenjoystories,Lamprinidis2023LLMCJ, Barkar2024, Chakrabarty2023ArtOA} resulted in score distributions that deviated from human-like patterns, raising questions about the reliability of LLMs in persuasiveness evaluation. \textbf{The first contribution of this paper is an original experimental protocol based on prompting the model to enhance or reduce persuasiveness in French public speech transcripts}. This novel approach allows us to assess the model's French-specific conceptual understanding of the persuasiveness concept instead of an old score-based persuasiveness evaluation approach.

Prior work has shown that texts generated by LLMs exhibit reduced lexical and syntactic diversity~\cite{Padmakumar2023DoesWW,guo2024curiousdeclinelinguisticdiversity,Chen2023STADEESD}. Also, creative writing experts pointed out that AI-generated creative texts struggle with narrative endings, language proficiency, literary devices, rhetorical complexity and unusual syntax~\cite{Chakrabarty2023ArtOA}. Unfortunately, such studies require dense and expensive human annotations difficult to collect for each context. We extend this literature approaches here by proposing a new methodology for automatically analysing the linguistic shifts in ChatGPT-enhanced text compared to the original human-produced speeches. Our method does not require an expert annotation, and our literature analysis proposes a highly informative summary of the lexical patterns exhibited in persuasive, clear and engaging texts or speeches. \textbf{The second contribution of this paper is the introduction of a set of new textual features\footnote{We provide an open-source implementation of the feature set for further research and replication on anonymous GitHub: \url{https://github.com/anonympapers/Persuasiveness_for_ChatGPT}} consisting of features used to model language in prior work \cite{features:Barkar, guo2024curiousdeclinelinguisticdiversity} extended with novel features inspired by rhetorical techniques described in~\cite{misner2023messages, forsyth2014elements}, such as storytelling indicators, alliteration, anaphora, etc.} In this work we focus on the GPT-4o model -- the latest and one of the most used by the population  for text writing and enhancement -- and focus on the following research questions:
 \\
 \textbf{RQ1:} How do textual features differ when prompting ChatGPT to enhance versus diminish the persuasiveness of speech transcripts?\\
\textbf{RQ2:} How do the differences in textual feature variations between enhanced and diminished texts align with the recommendations of public speaking experts and language patterns identified in the literature?

\section{Data}
\subsection{Overall Description}
\noindent \textbf{Motivations.} In our study, we utilise the 3MT\_French dataset, which comprises 3-minute presentations by French PhD students from the \textit{``Ma Thèse en 180 secondes''} competition. The dataset was selected based on the following criteria: (1) educational context, (2) consistent presentation duration, (3) availability as \textbf{the only open-source French} public speaking dataset, and (4) balance between non-professional speakers and well-prepared content, ensuring both scope for improvement and data quality.\\
\noindent \textbf{Data Description.} Data contains (\textit{135 female, 113 male}) presentations on diverse research topics. Each was rated on a 5-point Likert scale for persuasiveness with definitions adapted from the Public Speaking Competence Rubric~\cite{Schreiber2012TheDA}. For details on the annotation protocol, refer to~\cite{MT180}.

\subsection{Applied Post-processing}
\noindent \textbf{Transcription Quality Check.} The dataset does not include transcripts. We generated them using Whisper~\cite{whisper}, a state-of-the-art Automatic Speech Recognition model. After verifying and removing corrupted samples, 227 transcripts remained. 

\noindent \textbf{Quality 
 of Representativeness.} To enhance diversity in speech quality, we selected a balanced subset based on three criteria from human annotations of persuasiveness (the initial dataset was skewed toward scores of 3 or 4):
 \begin{itemize}
     \item at least one score of 1 or 2;
     \item at least two scores of 5;
     \item samples awarded a jury or public prize;
 \end{itemize}

 Then, for the samples chosen with the listed criteria we applied filtering. To ensure a uniform distribution of persuasiveness scores, we calculated the average number of samples per score within the chosen subset. This average, rounded to the nearest integer, defined the target sample size for each category (\texttt{mean\_count}). We then applied stratified sampling, drawing up to \texttt{mean\_count} samples from each score. This process balanced the representation of persuasiveness levels across the dataset, ensuring no score dominated. The final balanced subset was achieved by grouping on persuasiveness scores and sampling from each group with a fixed random seed for reproducibility. We used the harmonic mean to aggregate persuasiveness scores across raters, as it reduces the influence of outliers and emphasizes lower ratings. This conservative approach helps mitigate the effect of disproportionately high individual scores in public speaking evaluations. After this process, we ended up with 90 samples that form our 3MT\_French\_balanced\_subset.

\section{Methodology}

\subsection{Step 1: Generation of Upgraded and Downgraded Transcripts}\label{sec:generation}

Using the GPT-4o model via Open AI Playground API (historically utilized high-performance GPUs not publically disclosed), we generated enhanced (upgraded) and diminished (downgraded) versions of transcripts from the balanced subset of 3MT\_French. Our prompt specified the communicative context of texts (``Ma thèse en 180 secondes'' and provided a definition of persuasiveness from~\cite{MT180} in order to guide introduced changes\footnote{These prompts can be found on our GitHub}. Prompts also restricted outputted by GPT-4o texts to include new persuasiveness score, modified text\footnote{We also required to provide explanations of adjustments but this paper does not cover analysis of these explanations}. We note that the model systematically ($88\%$ samples) provided the upgraded version with a score of $4.5$ and downgraded with scores of $2$, $2.5$ or $3$ ($30-34\%$). Execution time and token use were 12 sec. with 1475 tokens on average for downgraded against 15 sec. and 1622 tokens for upgraded texts. After the generation, we ensured that outputs met requirements (\textit{i.e.} a new score was provided, new text was provided and was in French).

\subsection{Step 2: Textual Features Extraction}

In order to analyse linguistic shifts in upgraded and downgraded transcripts (Section \ref{sec:generation}), we extract textual features from all three versions. We rely on the previously defined features from \cite{guo2024curiousdeclinelinguisticdiversity, features:Barkar} and introduce new ones ($^{\star}$) that approximate numerically the use of rhetorical devices, text complexity and discourse navigation (in brackets we explain how the feature was calculated)\footnote{formulas can be found on our GitHub}. Additionally, we summarise each feature category's purpose in Table~\ref{tab:lexical_features}.
\\
\noindent \textbf{Discourse.} We quantify discourse with \textbf{overlap} features such as word repetition between sentences~\cite{features:Barkar}, \textbf{alliteration$^{\star}$} (count of repeated initial sounds(letter) for neighbour words), \textbf{anaphora$^{\star}$} (average length of repeated phrases at the start of clauses or sentences), \textbf{antimetabole$^{\star}$} (count of reversed phrase repetitions), and \textbf{epanalepsis$^{\star}$} (count of repetition of the first three words of a sentence within its last three). Another discourse quantifier is \textbf{transitions} features such as types, number and similarity of conjunctions~\cite{features:Barkar} extended with \textbf{expletives$^{\star}$} (mid-sentence fillers count~\textit{en fait, bien sûr, évidemment, etc.}), \textbf{polysyndeton$^{\star}$} (count of repetitive coordinating conjunctions within the sentence), and \textbf{asyndeton$^{\star}$} (count of conjunction omission in list-like structures identified through punctuation). New \textbf{overlap} and \textbf{transition} features approximate rhetorical techniques used in public speaking~\cite{misner2023messages, forsyth2014elements}. We introduce discourse \textbf{storytelling$^{\star}$} count-based features by extending discourse particles \textit{'alors', 'enfin', etc.}~\cite{Lee2019CanPM} with preliminary list of narrative indicators such as \textit{'il était une fois', 'raconter', etc.} and comparative indicators such as \textit{'comme', 'tel', 'pareil à', etc.}). Finally, we measure pronunciation complexity and listening difficulty of the discourse with \textbf{readability$^{\star}$}, represented by the Flesch reading-ease score previously used in video-advertising persuasiveness analysis~\cite{Komar2015}.
\\
\noindent \textbf{Lexical.} We used \textbf{lex\_ diversity}, measured as type-token ratio and measure of textual lexical diversity \cite{features:Barkar, guo2024curiousdeclinelinguisticdiversity}.
\\
\noindent \textbf{Syntactic.} We used \textbf{syn\_diversity} (Weisfeiler-Lehman isomorphism test for graph mapping of sentence tree \cite{guo2024curiousdeclinelinguisticdiversity}), \textbf{syn\_structures$^{\star}$} (proportions of clause, \textit{i.e.}, grammatical units such as declarative/interrogative/exclamatory), and \textbf{negation$^{\star}$} (occurrences of \textit{``ne..pas'', ``jamais'' etc}) features.
\\
\noindent \textbf{Psychological and Rhetorical.} Incorporates LIWC-based cognitive, affective and perceptive features~\cite{features:Barkar}.

\begin{table*}[!t]
\caption{Hand-crafted lexical features. New for public speaking features marked with $\star$.} \label{tab:lexical_features}
\centering
\resizebox{2\columnwidth}{!}{%
\begin{tabular}{|p{1.8cm}|p{3cm}|p{12.2cm}|}
    \hline
    \textbf{Complexity} & \textbf{Sub-Group} & \textbf{Description}\\ \hline

    \multirow{4}{*}{\makecell[l]{Discourse}} 

    & 7 Transitions & 4 measures similarity and occurrences of subordinating (e.g. \textit{quand, parce queue, etc.}) and coordinating (e.g. \textit{et, ou, mais, etc.}) conjunctions and were taken from \cite{features:Barkar}. 3 new~$\star$ serve as proxies for expletives, polysyndeton and asyndeton count inspired by \cite{misner2023messages, forsyth2014elements}.\\
    
    & 8 Overlap & 4 ratio of repeated words in different sentences (subsequent/in general) and were taken from \cite{features:Barkar}. 4 new~$\star$ serve as proxies for alliteration, anaphora, antimetabole, epanalepsis count inspired by \cite{misner2023messages, forsyth2014elements}. \\

    & 1 Readability Metrics $\star$ & Flesch reading-ease score measuring the difficulty of reading (the higher, the easier) inspired by \cite{Komar2015, Ta2022}.\\

    & 2 Storytelling $\star$ & Counts occurrences narrative indicators (e.g. \textit{'alors','bon', 'donc', 'enfin', 'quoi', 'voila'}) from \cite{Lee2019CanPM} extended with more narrative (e.g. \textit{'il était une fois',  'raconter', etc.}) and comparison indicators (e.g. \textit{'comme', 'tel', 'pareil à'}) indicators.  \\ 
     \hline


    \multirow{1}{*}{\makecell[l]{Lexical}} 
     & 2 Lexical Diversity & Measures the diversity of token types within segments of text with type-token ratio (TTR) and Measure of Textual Lexical Diversity (MTLD). Reduced list from \cite{essay,features:Barkar,guo2024curiousdeclinelinguisticdiversity}.\\
      \hline

    \multirow{3}{*}{\makecell[l]{Syntactic}}
    & 1 Syntactic Diversity& Measures the similarity of sentence structures within the text based on dependency trees and taken from \cite{guo2024curiousdeclinelinguisticdiversity}. \\
    & 1 Negation Count $\star$ & Counts occurrences of negation words/constructions (e.g., \textit{"ne..pas", "jamais"}). \\ 
    & 10 Syntactic Structure Count $\star$& Measure the percentage of different syntactic structures (e.g. \textit{declarative, relative, interrogative, exclamatory, passive voice etc.}). \\ \hline


    \multirow{4}{*}{\makecell[l]{Psy. \& Rhet.}} & &\multirow{4}{*}{\makecell[l]{Extracted using word count of Affective, Cognitive\\ and Perceptive processes \cite{LIWC}. Inspired by\\ \cite{teferra2023predicting, Ta2022, Guyer2021}}}\\
    
    & 6 Aff. LIWC &  \\
    & 9 Cog. LIWC & \\
    & 1 Prc. LIWC & \\
  
 \hline

\end{tabular}
}
\end{table*}

\section{Results \& Discussion}

\setlength{\tabcolsep}{3pt} 

\begin{table*}[t!]%
\centering%
\renewcommand{\arraystretch}{1.2}%
\caption{Averaged feature shifts between paired samples ($\uparrow$ for increased after generation values, $\downarrow$ for decreased). Italic indicates features that change in the same direction regardless of persuasiveness level, while bold marks features that shift differently when persuasiveness is upgraded or downgraded.}%
\resizebox{1.7\columnwidth}{!}{%
\large
\begin{tabular}{|p{2.5cm}|p{6.5cm}|p{6.5cm}|}%
\hline%
\textbf{Group} & \textbf{Initial $\rightarrow$ Upgraded} & \textbf{Initial $\rightarrow$ Downgraded} \\%
\hline%
\multicolumn{3}{|c|}{\textbf{Discourse}} \\%
\hline%
overlap & \textbf{local\_NOUN\_overlap}\(\downarrow\) 
\textit{global\_NOUN\_overlap}\(\downarrow\) 
\textit{global\_content\_overlap}\(\downarrow\) 
\textit{anaphora}\(\downarrow\) 
\textit{epanalepsis}\(\downarrow\)
& \textbf{local\_NOUN\_overlap}\(\uparrow\) 
\textbf{local\_content\_overlap}\(\uparrow\) 
\textit{global\_NOUN\_overlap}\(\downarrow\) 
\textit{global\_content\_overlap}\(\downarrow\) 
\textbf{alliteration}\(\uparrow\) 
\textit{anaphora}\(\downarrow\) 
\textit{antimetabole}\(\downarrow\) 
\textit{epanalepsis}\(\downarrow\)\\%
\hline%
transitions & \textit{transitions\_number}\(\downarrow\) 
\textit{trans\_similarity}\(\downarrow\) 
\textit{transition\_types\_count}\(\downarrow\) 
\textit{expletives\_count}\(\downarrow\) 
\textit{polysyndeton\_count}\(\downarrow\) 
\textit{asyndeton\_count}\(\downarrow\) 
& \textbf{transitions\_per\_sentence}\(\downarrow\) 
\textit{transitions\_number}\(\downarrow\) 
\textit{trans\_similarity}\(\downarrow\) 
\textit{transition\_types\_count}\(\downarrow\) 
\textit{expletives\_count}\(\downarrow\) 
\textit{polysyndeton\_count}\(\downarrow\) 
\textit{asyndeton\_count}\(\downarrow\)\\%
\hline%
storytelling & \textit{narrative\_words}\(\downarrow\) 
\textit{comparison\_count}\(\downarrow\) & 
\textit{narrative\_words}\(\downarrow\) 
\textit{comparison\_count}\(\downarrow\)\\%
\hline%
readability & \textit{flesch\_kincaid\_score}\(\downarrow\) 
& \textit{flesch\_kincaid\_score}\(\downarrow\)\\%
\hline%
\multicolumn{3}{|c|}{\textbf{Lexical}} \\%
\hline%
lex\_diversity & \textit{ttr}\(\uparrow\) \textit{mtld}\(\uparrow\) & \textit{ttr}\(\uparrow\) \textit{mtld}\(\uparrow\)\\%
\hline%
\multicolumn{3}{|c|}{\textbf{Syntactic}} \\%
\hline%
negation & \textit{negation\_count}\(\downarrow\) & \textit{negation\_count}\(\downarrow\)\\%
\hline%
syn\_diversity & \textit{syntactic\_diversity\_mean}\(\downarrow\) & \textit{syntactic\_diversity\_mean}\(\downarrow\)\\%
\hline%
syn\_structures & \textbf{Interrogative}\(\uparrow\) \textbf{Exclamatory}\(\uparrow\) \textbf{Imperative}\(\uparrow\) \textit{Passive}\(\uparrow\) \textit{Cleft}\(\downarrow\) \textit{Nominal}\(\downarrow\) 
& \textbf{Declarative}\(\uparrow\) \textbf{Interrogative}\(\downarrow\) \textbf{Exclamatory}\(\downarrow\) \textbf{Imperative}\(\downarrow\) \textit{Passive}\(\uparrow\) \textit{Cleft}\(\downarrow\) \textbf{Conditional}\(\uparrow\) \textit{Nominal}\(\downarrow\)\\%
\hline%
\multicolumn{3}{|c|}{\textbf{Psycholinguistic and Rhetorical}} \\%
\hline%
Aff\_LIWC & \textbf{affect (emotion)}\(\uparrow\) \textbf{positive\_emotion}\(\uparrow\) & \textbf{negative\_emotion}\(\downarrow\) \textbf{anxiety}\(\downarrow\) \textbf{anger}\(\downarrow\) \textbf{sadness}\(\downarrow\)\\%
\hline%
Cog\_LIWC & \textit{insight}\(\uparrow\) \textbf{causation}\(\downarrow\) 
& \textbf{cognition}\(\uparrow\) \textit{insight}\(\uparrow\) \textbf{tentative}\(\uparrow\) \textbf{certainty}\(\downarrow\) \textbf{exclusion}\(\downarrow\)\\%

\hline%
\end{tabular}%
}%
\label{tab:feature_comparison}%
\end{table*}%

In order to analyse the significance of shifts in linguistic features across transcript versions (initial vs. upgraded and initial vs. downgraded), we use the Mann-Whitney U Test, a non-parametric test for comparing two independent groups. It is applied pairwise (\textit{initial vs. upgraded and initial vs. downgraded}) to assess distribution differences. The test is performed with the two-sided alternative hypothesis, meaning it detects any significant difference without assuming a specific direction of change. We observed significant U-statistics (\textit{i.e.}, p-value $\leq 0.05$) for all groups of features, except for \textbf{Perc\_LIWC}. The most pronounced differences were observed in \textbf{lexical\_diversity}, while \textbf{syllable\_readability} displayed moderate variation. Other features exhibited distinct but less substantial changes. These findings suggest that  linguistic richness and structural complexity are notably altered in modified transcripts (\textbf{RQ1}). To assess feature shifts on the original scale and address \textbf{RQ1}, we computed average differences between paired samples (as  upgraded or downgraded each text) for features with significant Mann-Whitney U Test results and reported them in Table \ref{tab:feature_comparison}.

\subsection{ChatGPT-Style Linguistic Modifications (Independent of Persuasiveness Level).}

This section details systematic linguistic modifications in ChatGPT-generated texts, regardless of persuasiveness adjustment, as highlighted in Table~\ref{tab:feature_comparison}.

\noindent \textbf{Discourse Complexity.}  ChatGPT-generated texts exhibit significantly lower levels of content redundancy (\textbf{RQ1}), as reflected in the reduced statistics for \textbf{overlap} features. Specifically, ChatGPT minimizes global repetition and avoids anaphora (e.g., ``We will rise above challenges. We will shape the future.'') and epanalepsis (e.g., ``A leader must serve the people, for without the people, there is no leader.'').  These rhetorical devices, often used to enhance engagement and memorability in speech~\cite{misner2023messages, forsyth2014elements}, are notably suppressed (\textbf{RQ2}). Additionally,  ChatGPT-generated texts significantly reduce the use of comparisons and narrative indicators (\textbf{RQ2}) as evidenced by lower \textbf{storytelling} feature values. These elements serve as discourse particles~\cite{Lee2019CanPM} and help in structuring the discourse, expressing attitudes and emotions, steering the flow and emphasising the message~\cite{Stede2000,Schourup1983}. This may contribute to a more formal and structured style but potentially at the expense of expressiveness and audience involvement. In terms of readability,  ChatGPT-generated texts are consistently more difficult to read (RQ1) than the original transcripts. The Flesch-Kincaid Score decreased from an average of $64$ (``Plain text, easily understood by 13- to 15-year-old students'') in the initial speeches to approximately $47$–$48$ (``Difficult to read'') in generated texts. This trend contrasts with the findings of \cite{Komar2015} (\textbf{RQ2}), who observed that persuasive advertisements typically employ simpler language accessible to younger audiences between 8 and 13 y.o. However, while more complex texts may be harder to follow in spoken form, they may also engage audiences more deeply ~\cite{Ta2022} (\textbf{RQ2}). Finally, a  consistent reduction in transitions within and between sentences (\textbf{RQ1}) is observed, marked by lower frequencies of conjunctions and transition variety (\textit{i.e.}, transition\_number and transition\_similarity). Furthermore,  ChatGPT reduces the use of persuasive emphasis techniques, which contribute to rhetorical impact in public speaking (\textbf{RQ2}) as seen from decline of expletives (single-word emphasis), polysyndeton (repetitive conjunction use), and asyndeton (conjunction omission)~\cite{misner2023messages, forsyth2014elements}. This finding aligns with expert opinion on problems of LLMs with literary devices use and overall creative writing quality~\cite{Chakrabarty2023ArtOA}.

\noindent \textbf{Lexical Complexity.}  ChatGPT-generated texts consistently exhibit higher lexical diversity (\textbf{RQ1}), i.e. \textit{ttr}, \textit{mtld}. This contrasts with prior research~\cite{Ta2022}, which argues that lexical repetition enhances persuasion by simplifying lexical structures while maintaining complex relationships (\textbf{RQ2}). It aids in structuring arguments through textual markers that improve coherence and logical flow. Thus, lower lexical diversity contributes to more cohesive and persuasive communication.

\noindent \textbf{Syntactic Complexity.}  Across all generated texts, syntactic diversity is reduced (\textbf{RQ1}) as indicated by syntactic\_diversity\_mean. This suggests a simplification of sentence constructions, aligning with prior findings on the declining syntactic diversity of LLM-generated text~\cite{guo2024curiousdeclinelinguisticdiversity}. Additionally, ChatGPT exhibits a  notable reduction in negation constructions (\textbf{RQ1}), potentially contributing to a more positive textual tone.  Lower frequency of cleft constructions (\textit{i.e., c’est/ce qui})  and nominal sentences also suggest syntactic structure simplification (\textbf{RQ1}). Moreover,  passive voice usage increases across both transcript types (\textbf{RQ1}), which may contribute to perceptions of objectivity and clarity~\cite{Ruelas2020PassiveVoice} (\textbf{RQ2}).

\subsection{Persuasiveness-Dependent Modifications in ChatGPT-Generated Texts}

This section examines linguistic features that differ between upgraded and downgraded transcripts, as highlighted in bold in Table \ref{tab:feature_comparison}.

\noindent \textbf{Discourse Complexity.} Interestingly,  downgraded speeches retain higher levels of local repetition and alliteration (\textbf{RQ1}), which contributes to rhythmic and stylistic variation (\textbf{RQ2}). However, they exhibit a marked reduction in an antimetabole, a technique known to enhance persuasiveness by reinforcing key phrases through reversal~\cite{misner2023messages, forsyth2014elements} (\textbf{RQ2}).  This suggests that less persuasive speeches generated by ChatGPT may lack the structural reinforcement typically found in impactful public discourse (\textbf{RQ2}).

\noindent \textbf{Syntactic Complexity.}  Upgraded transcripts exhibit an increased use of interrogative, exclamatory, and imperative sentences (\textbf{RQ1}) that found to ask rhetorical questions~\cite{misner2023messages, forsyth2014elements}, capture attention~\cite{Komar2015}, and foster a direct connection with listeners~\cite{Komar2015} respectively (\textbf{RQ2}). This diversity of clauses aligns with \cite{Guyer2021}, who found that a mix of exclamatory and interrogative intonations enhances confidence and engagement (\textbf{RQ2}). Conversely,  downgraded transcripts favour declarative and conditional constructions, suggesting a more neutral and less emotionally charged delivery (\textbf{RQ1}). This trend may contradict \cite{Ta2022} (\textbf{RQ2}), who found that lower emotional intensity can sometimes enhance persuasion, depending on context.

\noindent \textbf{Psychological and Rhetorical Complexity.} The analysis revealed  distinct patterns in affective and cognitive lexicons expressed with Aff\_LIWC and Cog\_LIWC across upgraded and downgraded transcripts (\textbf{RQ1}). Upgraded transcripts demonstrate greater use of affective (\textit{e.g., heureux, triste, amour, détester}) and positive emotion (e.g., \textit{joyeux, satisfait}) terms, suggesting that  ChatGPT enhances persuasiveness by increasing emotional expressiveness (\textbf{RQ1}). Downgraded transcripts, on the other hand, contain fewer emotion-related words (\textit{i.e. emoneg}) and avoid anxiety, anger, and sadness terms, indicating that  low persuasiveness is associated with a more neutral emotional tone (\textbf{RQ1}). The literature presents a nuanced view on this phenomenon (\textbf{RQ2}): \cite{East2007} notice that people have general tendency to use positive words of mouth more frequently than negative; \cite{Ho2009} argues that emotional expression is a strategic choice and can be used to increase persuasiveness; \cite{Ta2022} report that messages marked with less emotionality had higher odds of persuasion than messages marked with more emotionality, regardless of whether it was positive or negative. However, more in-depth analysis of persuasion in dialogues~\cite{Tan_2016} argues towards positive emotionality (also measures with LIWC lexicon) in persuasive messages. For cognitive lexicons,  upgraded transcripts reduce causal markers (\textbf{RQ1}) (e.g., \textit{parce que, donc, cependant}), while  downgraded speeches feature more cognition-related verbs (e.g., \textit{penser, savoir, croire}),  tentative expressions (e.g., \textit{peut-être, probablement}), and  a lower frequency of exclusionary constructions and certainty markers.  This shift reduces the authoritative tone in downgraded transcripts, which may impact their persuasiveness (\textbf{RQ1}).

\subsection{Persuasion Theory Behind ChatGPT's Textual Enhancement}

One might argue that our findings are influenced by the distinction between spoken (initial) and written (generated) speech and not by the persuasion strategy preferred by ChatGPT. However, the increased use of emotional language in the upgraded transcripts (despite their written nature) contrasts with the typically lower emotional expressiveness of human-written texts~\cite{Berger2021}. Moreover, in our study, we not only specify the context of spoken performance when prompting the model, but also use originally prepared speeches, most probably carefully written by the speakers, and therefore closer resemble written content. Meanwhile, some observed patterns may have been chosen due to the written nature of generated speeches, for example, the higher detachment observed in generated transcripts (e.g., increased passive voice) aligns with previous findings on comparison between spoken and written human language~\cite{Redeker1984}, where authors observed avoidance of personal involvement in written texts.

Our results suggest that, while adapting to a public speaking context, ChatGPT does not replicate human-like persuasive strategies. Its generalised linguistic ``style'' aligns with the peripheral-route persuasion strategy in Elaboration Likelihood Model~\cite{petty1986communication} (\textbf{RQ2}), where emotional and stylistic manipulations appeal over a nuanced balance of logical argumentation, rhetorical devices, and structural creativity. This interpretation of ChatGPT persuasiveness manipulations is further supported by the Heuristic-Systematic Model (HSM)~\cite{Chaiken1980HeuristicVS}, which similarly distinguishes between systematic processing (effortful, message-focused) and heuristic processing (based on superficial cues such as source likability or emotional tone). Our findings indicate that ChatGPT favours heuristic-style construction: persuasive upgrades include more stylistic markers (e.g., imperative mood, affective language) and fewer message-level arguments or contrasts. According to both models, such emotional manipulation strategies are more persuasive under low-involvement conditions (HSM), but the resulting opinions are typically less stable or durable over time (HSM, ELM). This aligns with the idea that ChatGPT-generated upgrades enhance surface appeal rather than substantive argument strength. Furthermore, the increasing positive emotions in the upgraded transcripts align with the value-based decision-making framework~\cite{Falk2018Persuasion} that offers a neurocognitive perspective on the persuasion process. Specifically, this model conceptualises persuasion as a valuation process involving self- and social-relevance computations within the brain’s reward systems (e.g., ventromedial prefrontal cortex, ventral striatum). They discuss that activity in the ventromedial prefrontal cortex was greater for gain-framed messages (i.e. towards positive outcomes) than for loss-framed messages. However, also emphasise the importance of other factors such as value the receiver assigns to the message or action, synchrony in neural activity between communicators and receivers, and social relevance.

Taken together, these models provide converging evidence that ChatGPT's persuasive modifications operate predominantly through affective and stylistic routes rather than logical or dialogic reasoning. While this approach may increase initial message appeal, it likely limits long-term opinion change, perceived authenticity, or resistance to counterargument. The observed pattern suggests that ChatGPT’s persuasive strategy relies more on shallow cue-based enhancement than on deeply elaborated, context-sensitive communication.

\section{Conclusion}



When answering \textbf{RQ1}, we observed that regardless of persuasiveness level ChatGPT exhibits systematic stylistic modifications characterized by increased lexical diversity and reduced syntactic and discourse complexity. ChatGPT enhanced persuasiveness by increasing positive emotions, cognitive language, and favouring interrogative and imperative structures, while diminishing it with emotionally neutral language, with declarative and conditional constructions. By comparing to the literature on public speaking to address \textbf{RQ2}, we observed that emotional manipulations align with the short-term persuasiveness strategy from ELM~\cite{petty1986communication} and heuristic processing from HSM~\cite{Chaiken1980HeuristicVS}, while its tendency to avoid rhetorical devices contradicts traditional emphasis techniques~\cite{misner2023messages, forsyth2014elements}. Future research should explore how fine-tuning models on nuanced persuasive strategies could improve their rhetorical effectiveness. The new methodology can be also adapted to study LLMs' understanding of the cultural impact, audience and context adaptation on public speech. 
\section*{Limitations}

Despite the novel contributions and promising potential of this study, several important limitations must be acknowledged when interpreting the findings.

First, the dataset used in this study consisted of only 90 paired speech samples. Although a deliberate data selection procedure was applied to ensure a more balanced distribution across performance quality levels—avoiding overrepresentation of highly persuasive speeches—the sample size remains limited. As such, caution should be exercised when attempting to generalize the findings to broader public speaking contexts.

Second, the scope of the study was restricted to French-language speeches, as the overarching goal was to advance research on French public speaking performance. While this focus is justified, it limits the applicability of the results to other languages. Given that languages vary significantly in syntactic complexity and word order (e.g., German, Russian, or Arabic), further studies are needed to evaluate whether the observed trends hold across typologically different languages.

Third, although the study leveraged a comprehensive set of rule-based lexical features to approximate rhetorical devices, such methods have intrinsic limitations. These features were designed to serve as proxies for rhetorical constructions; however, not all true rhetorical devices may have been captured, and some detected patterns may not correspond to genuine rhetorical usage. A deeper investigation, ideally supported by detailed human annotations (e.g. human-annotated metaphors or storytelling), is necessary to validate these approximations. While such annotations are time-consuming and resource-intensive, they could significantly enhance the reliability of rhetorical feature analysis.

Finally, the present study examined only a single large language model—GPT-4o. Given the rapid development of new and diverse LLM architectures, it is currently impractical to perform in-depth comparative analyses across multiple models within a single paper. Nonetheless, we believe that the methodological framework and literature-informed feature set proposed in this study can serve as a foundation for evaluating other flagship models in future research. Such extensions will help determine whether the observed linguistic and rhetorical shifts are specific to GPT-4o or generalizable across different LLMs.

\bibliography{custom}
\bibliographystyle{acl_natbib}



\end{document}